\documentclass[letterpaper, 10 pt, conference]{ieeeconf}  

\IEEEoverridecommandlockouts                              

\usepackage{graphicx}
\usepackage{graphics} 
\usepackage{mathptmx} 
\usepackage{times} 
\usepackage{amsmath} 
\usepackage{amssymb}  
\usepackage{flushend}

\usepackage{graphicx}
\usepackage{caption}
\usepackage{subcaption}

\usepackage[linesnumbered,ruled,vlined,algo2e]{algorithm2e}
\SetKwRepeat{Do}{do}{while} 
\usepackage{dirtytalk} 
\usepackage{fancybox} 

\usepackage{url}

\usepackage{color} 

\title{\LARGE \bf
Improving the Modularity of  AUV Control Systems \\using Behaviour Trees
}

\author{Christopher Iliffe Sprague, \"Ozer \"Ozkahraman, Andrea Munafo, \\ Rachel Marlow, Alexander Phillips, Petter \"Ogren
         \thanks{The authors are with the Robotics, Perception and Learning Lab. (RPL), School of Electrical Engineering and Computer Science, Royal Institute of Technology (KTH), SE-100 44 Stockholm, Sweden, and with the Marine Autonomous and Robotic Systems at the National Oceanography Centre, Southampton, SO14 3ZH, United Kingdom.
e-mail: \tt{ $\{$sprague$|$ozero$|$petter$\}$@kth.se}
\tt{ $\{$andmun$|$raclow$|$abp$\}$@noc.ac.uk}
}}

\begin{document}
\maketitle
\thispagestyle{empty}
\pagestyle{empty}

\begin{abstract}
In this paper, we show how behaviour trees (BTs) can be used to design modular, versatile, and robust control architectures for mission-critical systems. In particular, we show this in the context of autonomous underwater vehicles (AUVs). 
Robustness, in terms of system safety, is important since manual recovery
of AUVs is often extremely difficult.
Further more, versatility is important to be able to execute many different kinds of missions.
Finally, modularity is needed to achieve a combination of robustness and versatility, as the
complexity of  a versatile systems needs to be encapsulated in modules, in order to create a simple overall structure enabling robustness analysis.
The proposed design is illustrated using a typical AUV mission.

\end{abstract}



\section{Introduction}
\label{sec:introduction}

\begin{figure}[t]
  \centering
    \includegraphics[width=0.99\columnwidth]{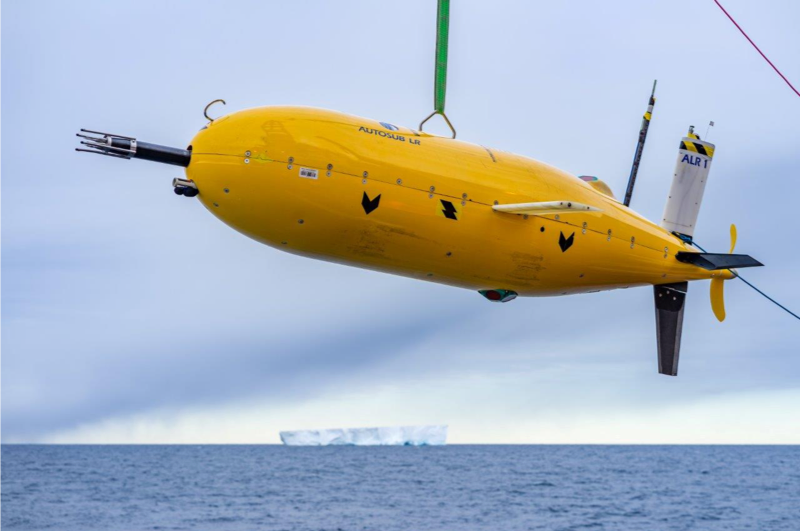}
  \caption{The National Oceanography Centre's Autosub Long Range 6000 AUV, prior to deployment under the Filchner-Ronne Ice Shelf in Antarctica. }
  \label{fig:NERC}
\end{figure}

Due to the limitation in communication bandwidth and range, Autonomous Underwater Vehicles (AUVs), such as the one in Figure~\ref{fig:NERC}, cannot rely on tele-operation solutions in the same way that aerial and ground-based robots do~\cite{fong2001vehicle}.
Furthermore, the areas of operation of AUVs are often remote and hard to access by other means,
 making recovery of a malfunctioning AUV very difficult and costly \cite{McPhail:2009up}.
For these reasons, the AUV control system needs to be both robust, to prevent the loss of vehicles,
 and versatile, to autonomously handle many different situations, see e.g., \cite{ozkahraman20183d}, on its own.
 These requirements lead to a trade-off in the complexity of the system,
 as robustness is often achieved by simplicity while versatility leads to complex systems.
 With this contribution we show how Behaviour Trees (BTs) can be used to design highly modular, versatile as well as robust, control architectures for AUVs.

BTs were invented in the computer game industry as a way to design complex, but modular, AI
for in-game opponents \cite{mateas2002abl,isla2005handling,champandard2007understanding}.
Since then, they have been shown to generalise a significant set of popular robotic control architectures, including Finite State Machines, Decision Trees, the Subsumption Architecture and the Teleo-reactive approach \cite{colledanchise2016behavior,colledanchise2017behavior}.
BTs have also been successfully used in the context of machine learning \cite{Nicolau2016,colledanchise2018learning,sprague2018adding}, 
training by demonstration \cite{olivenza2017trained},
collaborative assembly robotics, \cite{paxton2016costar}
and surgical robotics \cite{hu2015ablation}.

\section{Background}
\label{sec:classicalBT}
It this section we give a brief description of BTs, but refer the
interested reader to \cite{colledanchise2018book} for more details.

At the core, BTs are built from a small set of simple components, but just as many other powerful concepts,  
these components can be combined to create very rich structures.

Formally speaking, a BT is an acyclic directed graph (a tree) composed of nodes and edges.
The nodes of the tree can be divided into six categories, 
\emph{Fallback}, 
\emph{Sequence}, 
\emph{Parallel}, 
\emph{Action}, 
\emph{Condition}, 
\emph{Decorator},
as listed in Table \ref{tab:nodeTable}.

With the standard meaning of child and parent in the tree context,
the non-parent nodes are the interfaces to the physical AUV in terms of either
\emph{Actions}, where typically the AUV actuators are given commands, or
\emph{Condition}, where typically a condition is checked based on sensor data. 
All parent nodes of the tree are of the types \emph{Fallback}, \emph{Sequence}, \emph{Parallel}, 
and \emph{Decorator},
and are  used to decide what \emph{Actions} to execute,
based on both \emph{Conditions} and the outcomes of other \emph{Actions}.
A BT is executed by regularly sending \emph{Ticks} to the root node. This is done with a user defined frequency, chosen depending on how fast the system should be able to react and how much computing power is available. When a node is ticked, it either forwards the tick to one of its children, as explained below, or does some type of computation, typically involving sensor data if it is a \emph{condition}, or the transmission of commands to the actuators of the system if it is an \emph{action}.
We will now describe all node types in more detail.

\begin{figure}[ht]
\centering
\includegraphics[width=0.3\columnwidth]{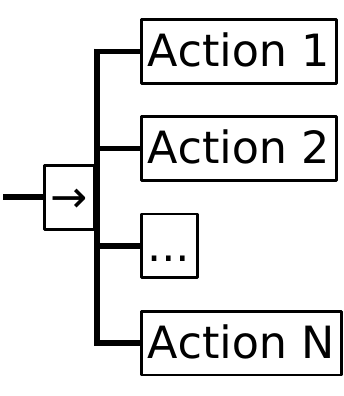}
\caption{Graphical representation of a Sequence node with $N$ children.}
\label{bts.fig.seq}
\end{figure}

The Sequence node executes Algorithm \ref{bts:alg:sequence},
which corresponds to routing the Ticks to its children from the left until it finds a child that returns either \emph{Failure} or \emph{Running}, then it returns \emph{Failure} or \emph{Running} accordingly to its own parent. It returns \emph{Success} if and only if all its children return \emph{Success}. Note that when a child returns \emph{Running} or \emph{Failure}, the Sequence node does not route the Ticks to the next child (if any). The Sequence node is drawn using the symbol \say{$\rightarrow$}, as shown in Figure~\ref{bts.fig.seq}. 

\begin{algorithm2e}[ht]
\SetKwProg{Fn}{Function}{}{}

\Fn{Tick()}
{
  \For{$i \gets 1$ \KwSty{to} $N$}
  {
    \ArgSty{childStatus} $\gets$ \ArgSty{child($i$)}.\FuncSty{Tick()}\\
    \uIf{\ArgSty{childStatus} $=$ \ArgSty{Running}}
    {
      \Return{Running}
    }
    \ElseIf{\ArgSty{childStatus} $=$ \ArgSty{Failure}}
    {
      \Return{Failure}
    }
  }
  \Return{Success}
  }
  \caption{Pseudocode of a Sequence node with $N$ children}
  \label{bts:alg:sequence}
\end{algorithm2e}

The Fallback node\footnote{Fallback nodes are sometimes also called \emph{selector} or \emph{priority selector} nodes.} executes Algorithm \ref{bts:alg:fallback},
which corresponds to routing the Ticks to its children from the left until it finds a child that returns either \emph{Success} or \emph{Running}. It then returns, accordingly, \emph{Success} or \emph{Running} to its own parent. It returns \emph{Failure} if and only if all its children return \emph{Failure}. Note that when a child returns \emph{Running} or \emph{Success}, the Fallback node does not route the Ticks to the next child (if any). 
The Fallback node is drawn using the symbol \say{?}, as shown in Figure~\ref{bts.fig.seq}. 

\index{Fallback node}
\begin{figure}[ht]
\centering
\includegraphics[width=0.3\columnwidth]{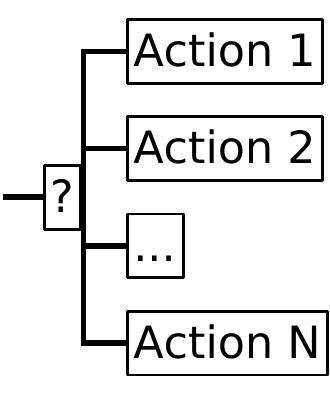}
\caption{Graphical representation of a Fallback node with $N$ children.}
\label{bts.fig.sel}
\end{figure}

\begin{algorithm2e}[ht]
\SetKwProg{Fn}{Function}{}{}

\Fn{Tick()}
{
  \For{$i \gets 1$ \KwSty{to} $N$}
  {
    \ArgSty{childStatus} $\gets$ \ArgSty{child($i$)}.\FuncSty{Tick()}\\
    \uIf{\ArgSty{childStatus} $=$ \ArgSty{Running}}
    {
      \Return{Running}
    }
    \ElseIf{\ArgSty{childStatus} $=$ \ArgSty{Success}}
    {
      \Return{Success}
    }
  }
  \Return{Failure}
  }
  \caption{Pseudocode of a Fallback node with $N$ children}
    \label{bts:alg:fallback}
\end{algorithm2e}

The Parallel node executes Algorithm \ref{bts:alg:parallel},
which corresponds to ticking all of its children. The return status is then given by an
aggregate of the return status of the children. Given a parameter $M\leq N$, where $N$ 
is the number of children, it returns \emph{Success} if at least $M$ children have succeeded,
\emph{Failure} if enough have failed to make the above impossible, and \emph{Running} otherwise.


\begin{algorithm2e}[ht]
\SetKwProg{Fn}{Function}{}{}

\Fn{Tick()}
{
  \ForAll{$i \gets 1$ \KwSty{to} $N$}
  {
    \ArgSty{childStatus}[i] $\gets$ \ArgSty{child($i$)}.\FuncSty{Tick()}\\
    }
    \uIf{$\Sigma_{i: \ArgSty{childStatus}[i]=Success}1 = N$}
    {
      \Return{Success}
    }
    \ElseIf{$\Sigma_{i: \ArgSty{childStatus}[i] =Failure}1 > 0$}
    {
      \Return{Failure}
    
  }\Else{
  \Return{Running}
  }
  }
    \caption{Pseudocode of a Parallel node with $N$ children and success threshold $M$}
  \label{bts:alg:parallel}
\end{algorithm2e}


\begin{algorithm2e}[ht]
\SetKwProg{Fn}{Function}{}{}

\Fn{Tick()}
{
 \ArgSty{ExecuteCommand()} \\
    \uIf{action-succeeded}
    {
      \Return{Success}
    }
    \ElseIf{action-failed}
    {
      \Return{Failure}
    }
    \Else
    {
    \Return{Running}
    }
   }
  \caption{Pseudocode of an Action node}
  \label{bts:alg:action}
\end{algorithm2e}

\begin{algorithm2e}[ht]
\SetKwProg{Fn}{Function}{}{}

\Fn{Tick()}
{
    \uIf{condition-true}
    {
      \Return{Success}
    }
    \Else
    {
      \Return{Failure}
    }
   }
  \caption{Pseudocode of a Condition node}
  \label{bts:alg:condition}
\end{algorithm2e}

When an Action node receives Ticks, it executes a command, as in Algorithm~\ref{bts:alg:action}. It returns \emph{Success} if the action is successfully completed or \emph{Failure} if the action has failed. While the action is ongoing it returns \emph{Running}.

When a Condition node receives Ticks, it checks a condition, as in Algorithm~\ref{bts:alg:condition}. It returns \emph{Success} or \emph{Failure} depending on if the condition holds or not. Note that a Condition node never returns the status  \emph{Running}. 

The Decorator node is a control flow node with a single child that manipulates the return status of its child according to a user-defined rule and also selectively Ticks the child according to some predefined rule. For example,
a \emph{Max 10 sec} decorator can start an internal timer the first time it is ticked, it then ticks its child and returns the corresponding return status for up to 10 seconds. After that it does not tick its child, but instead directly returns \emph{Failure}, thus effectively limiting the execution time of its child to 10 seconds. Other examples of decorators are the 
\emph{invert} decorator that always ticks its child, but swaps \emph{Success}/\emph{Failure} in the return status. 
For a more thorough description of BTs we refer the reader to the references given above.

\begin{table}[ht]
\scriptsize
\begin{center}
\begin{tabular}{|c|c|c|c|c|c|}
\hline
 \bf{Node type} & \bf{Symb.}& \bf{Succeeds if} & \bf{Fails if} & \bf{Running if} \cr

\hline
 Fallback  &\fbox{?} &  one child succ. &  all ch. fail & 1  ret. Running \cr
\hline
Sequence &\fbox{$\rightarrow$}&  all ch. succ. &  one ch. fails & 1  ret. Running \cr
\hline
Parallel &\fbox{$\rightrightarrows$} &  $\geq M$ ch. succ. &  $>N-M$ ch. fail &else \cr
\hline
Action & \fbox{text}& Succeeded &  Failed & else \cr
\hline
Condition & \ovalbox{text} & If true & If false & Never  \cr
 \hline
Decorator & $\Diamond$& Custom  & Custom & Custom \cr
\hline
\end{tabular}
\end{center}
\caption{The node types of a BT. See the detailed description for an explanation of the abbreviations used.}
\label{tab:nodeTable}
\end{table}%

\section{Problem Formulation}
\label{sec:problem}

\begin{figure}[!h]
  \centering
  	\includegraphics[width=0.99\columnwidth]{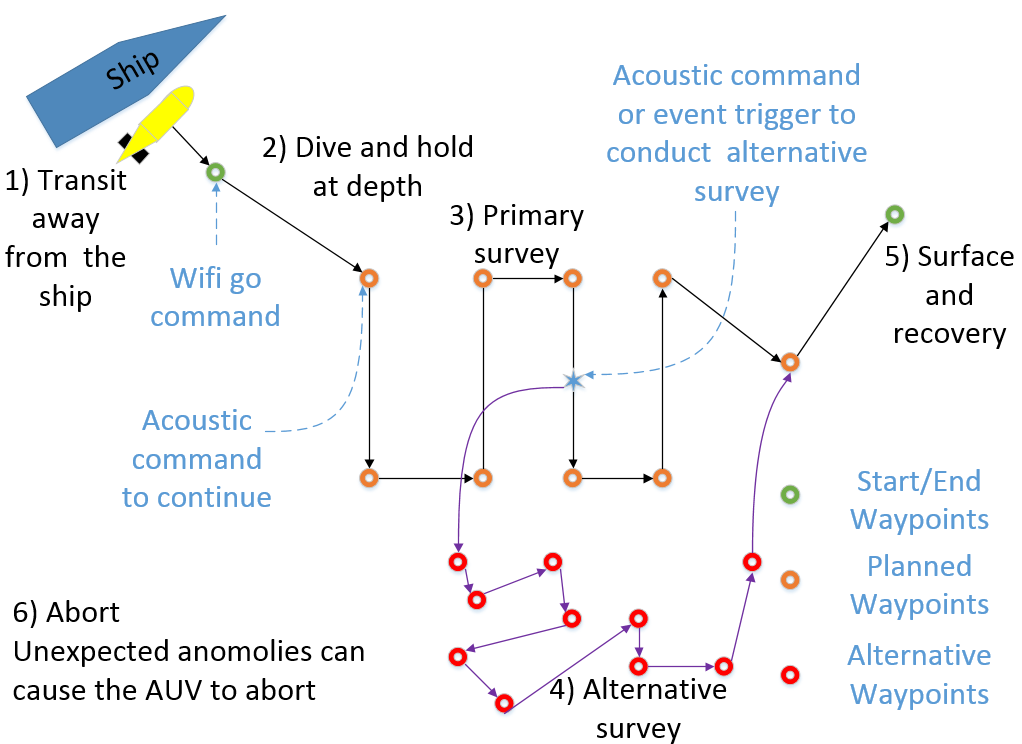}
  \caption{A typical AUV survey mission. }
  \label{fig:mission}
\end{figure}

A typical AUV deployment is illustrated in Figure~\ref{fig:mission}, and such missions are indicative of ship based deployments of AUVs \cite{DynOPO}. The example assumes Wi-Fi and/or Iridum communication between the operator and the AUV while it is on the surface. Whilst submerged we assume low bandwidth acoustic communications enabling the operator to upload simple commands or new mission tasks (e.g., waypoints) and download basic diagnostic information. In this example the mission consists of six possible phases: 
\begin{itemize}
    \item \emph{Transit away from the ship -} initial phase where the AUV moves to a safe stand-off distance from the launch vessel and waits for an operator go command prior to diving.
    \item \emph{Dive and hold at depth} - the AUV dives, conducts in water calibration(s) (e.g. compass) and turns on payload(s) transits to a target depth and location while being monitored acoustically. The AUV waits for an acoustic command to continue.
    \item \emph{Primary survey} - the vehicle navigates between a series of waypoints sequentially, often resulting in a lawnmower pattern, while maintaining a constant water depth or altitude from the seabed.
    \item \emph{Alternative survey} - triggered either by an acoustic operator command or by a sensor reading the AUV may be tasked to undertake a secondary survey;
    \item \emph{Surface and recovery} - the final mission phase where the AUV returns to the surface, turns off payloads and is recovered from the water.
    \item \emph{Abort} in the event of unexpected anomalies (e.g. a detected leak) the vehicle will undertake an abort behaviour. In open ocean this involves dropping a weight while the AUV attempts to surface. When operating under ice more complex contingency behaviours are required. Such anomalous events include: over depth condition, under altitude condition, leak detected, propeller stuck detected, actuator stuck detected etc.
\end{itemize}

With the example mission structure in mind, we now consider what properties we want the AUV control system to exhibit. In general, the system should try to satisfy the following properties, or exhibit the corresponding behaviours, in the given order of priority. The AUV should:
\begin{enumerate}
\item \textit{Maintain the safety of the AUV} - the AUV should abort the mission in case of high risk anomalies (e.g. leaks, propeller stuck etc.) but also avoid obstacles appearing in its path and maintain a safe clearance from the seabed.
\item \textit{Wait for operator clearance} - specific phases of the mission should only start following a command from the operator. E.g. the primary survey should only start on receipt of a \textit{continue} acoustic command from the operator indicating the system is fully prepared (e.g. payloads on and functioning correctly). This is particularly relevant for high risk operations e.g. under ice or in deep water where initial monitoring of the system is recommended to mitigate the risk of vehicle loss \cite{brito2010risk}.
\item \textit{Keep Mission tasks synchronised} - the mission tasks (e.g. survey route) is updated in relation to the objectives of the operators and/or acquired sensor data \cite{Ferri:2018jla, Munafo:2017kg}. If the operator requires a task update, communications is necessary \cite{Caiti:2013fm}.
\item \textit{Finalise the Mission} - return to the surface, turn payload(s) off and await recovery.
\end{enumerate}

Considering the system's properties, two sets of overall conditions and actions, as well as their pairings (behaviours), are generated, which will form the building blocks with which the eventual BT will be built. We will see that, although some conditions and actions are repeated among the set of behaviours, it will not result in redundant code, as a key practical benefit of using behaviour trees is their accommodation of reusing actions in several places in the tree.

\section{Proposed Solution}
\label{sec:solution}

\begin{figure}[!h]
\centering
\includegraphics[width=\columnwidth]{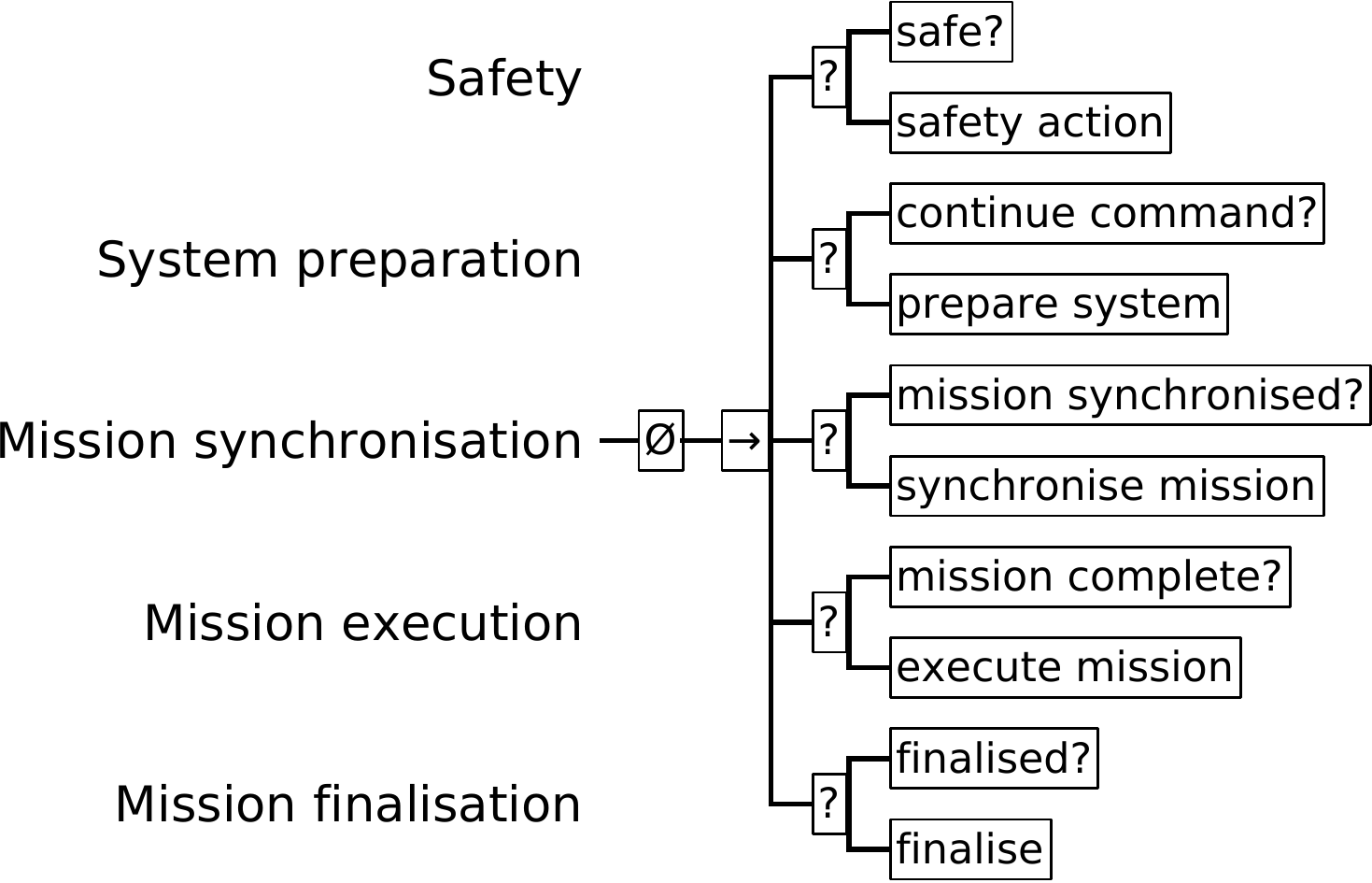} 
\caption{General mission critical system BT. Note the modular structure that will be expanded in Figure~\ref{fig:sbt}. Robustness is achieved by always checking the safety condition and taking appropriate actions when needed.}
\label{fig:gbt}
\end{figure}

Given the general mission properties outlined above, 
we create the general BT shown in Figure~\ref{fig:gbt}.
As can be seen, the BT closely follows the list of desired 
properties of the system. First it checks is the system is safe,
if not it performs a safety action. If safe, it goes on to check
if a continue command has been issued, if not it executes the system preparations. If ordered to continue, it checks if the mission is synchronised with received updates and sensor data, if not it updates its mission. If synchronised, it checks if the mission is complete, if not it goes on to execute the mission. If the mission is complete it finalises the mission e.g., by turning off payloads and returning to the surface. 

Note that all these checks are carried out with the frequency of the BT, making the system react immediately to a non safe situation, or new waypoints being received.
This construction of combining a condition with the action needed to satisfy the condition is a simple version of the the postcondition-precondition-action (PPA) structure, described in  \cite{colledanchise2018book}.



Now we can refine the general BT given in Figure~\ref{fig:gbt} to capture the details of the mission outlined in Figure~\ref{fig:mission}. We want to add the following details to the BT:
\begin{enumerate}
\item Safety: Abort mission in case of: inability to ascend/descend, actuator or propeller problems, leaks, too deep. Avoid obstacles on the way to the next waypoint.
\item System preparation: keep away from ship,  calibrate compass, payload on, go to target depth, wait for continue command.
\item Mission synchronisation: go to surface if commanded, update waypoints if commanded acoustically by the operator or if updated by an on-board algorithm monitoring sensor data.
\item Mission execution: go to next way point until all waypoints are visited.
\item Mission finalisation: turn payload off and go to surface
\end{enumerate}

\begin{figure}[!t] 
\centering
\includegraphics[width=0.98\columnwidth]{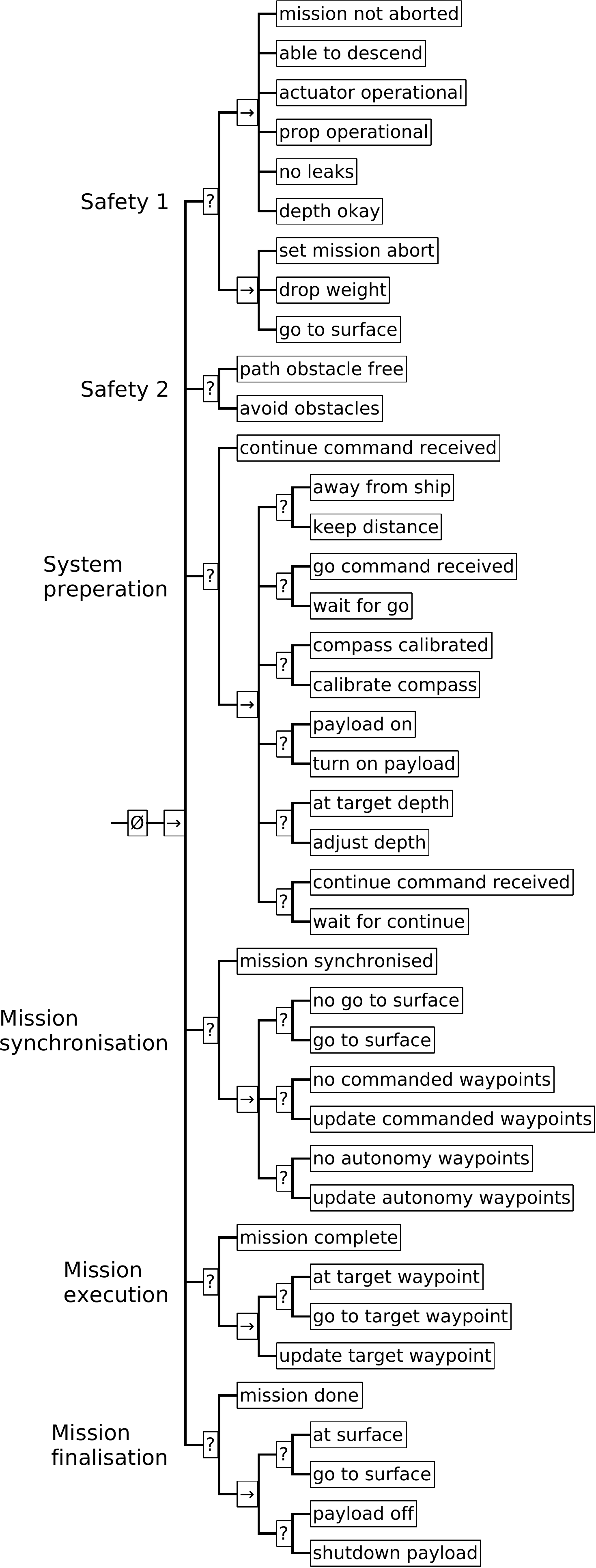}
\caption{AUV mission BT with the same overall structure as  Figure~\ref{fig:gbt}.}
\label{fig:sbt}
\end{figure}

The corresponding design can be seen in Figure~\ref{fig:sbt}.~
Note that the overall structure of Figure~\ref{fig:gbt} has been
maintained, with conditions being checked, and actions to satisfy those conditions being executed when needed, keeping the modular reactive design, while adding details to the execution.
Thus, the robustness of a well tested safety behaviour is not jeopardised by adding complex details to the mission execution sub-tree.
One could thus keep expanding upon the BT to an arbitrary level of detail.

As seen above, the modularity of BTs is very convenient when manually designing a solution given a set of requirements.
However, this modularity is also very important for automatic synthesis of BTs using planning tools~\cite{colledanchise2016planning} or machine learning~\cite{colledanchise2018learning,sprague2018adding}.

\section{Example Executions}
\label{sec:example}


To illustrate the functionality of the proposed BT, we describe some sample mission executions.
Suppose the AUV is submerged in proximity to the ship, as would be expected in the nominal mission scenario; this would yield the following condition checks and action executions: (but note that each number in the list below can correspond to one or arbitrarily many ticks, depending on how long the operation takes)
\begin{enumerate}
\item Mission not aborted, able to ascend/descend, actuator and prop are operational, there aren't any leaks, and not over depth, so don't drop weight and go to surface (note that this condition is continuously checked).
\item Not away from ship, so transit away from ship.
\item Now away from ship, wait for go command.
\item Go command received, so stop waiting!
\item Is the compass calibrated? No, so calibrate it!
\item Compass is now calibrated, check if the payload is powered on.
\item Payload is now powered on, check if at target depth.
\item Not at target depth, so go to target depth.
\item Now at target depth, wait for the continue commands.
\item Received continue command, so stop waiting!
\item No commands to surface, so don't surface.
\item No user commanded waypoints, so don't update waypoints.
\item No waypoints received from autonomy, so don't update way points.
\item All waypoints not visited, so check position.
\item Not at target waypoint, so go to target way point.
\item Now at target waypoint, so assign new target as next way point.
\item Not at the new target waypoint, so go to it.
\item New waypoints received from either the operator or on-board autonomy, so updated waypoints.
\item Keep following new waypoints.
\item Eventually at terminal waypoint (mission complete), check if at surface.
\item Not at surface, so go to surface.
\item Now at surface, so check if payload is off.
\item Payload is not off, so shut it down.
\item Mission is now finalised.
\end{enumerate}

Now consider a scenario in which a critical condition is encountered during the course of the mission execution:
\begin{enumerate}
\item Mission not aborted, able to ascend/descend, actuator and prop are operational, there aren't any leaks, and not over depth, so don't drop weight and go to surface.
\item ...
\setcounter{enumi}{14}
\item Not at target waypoint, so go to target way point.

\item Still able to ascend/descend, actuator and prop are operational, but there is a leak! Set Mission Abort to TRUE. Drop weight and go to surface.
\item Leak might stop due to reduced pressure, but Mission abort is still TRUE, so continue to surface.
\item Stay at surface until picked up.
\end{enumerate}
As can be seen above, the proposed controller results in a reactive and fault-tolerant mission execution. Furthermore, the resulting AUV mission specific BT retains the high level functionality described by the general mission critical system BT in Figure~\ref{fig:gbt}.

\section{Conclusion and discussion}
\label{sec:conclusion}
In this paper we have demonstrated that the BT framework is an advantageous approach for controlling mission critical systems. We have shown the sequence of steps one can take in order to transcribe the given system and mission requirements in a succinct, modular, and convenient manner, proceeding from a general design as shown in Figure~\ref{fig:gbt} to a specific design in Figure~\ref{fig:sbt}.

With respect to safety, we have demonstrated that we are able to invoke safety checks with high precedence using node priority. We are able to ensure that at every moment in time the BT  monitors the system's safety before proceeding to other subsequent tasks.

With respect to modularity, we have demonstrated that, although there may be several behaviours that share mutual conditions and/or actions, single nodes can be implemented in a decentralised manner among several behaviours, eliminating the need for unnecessary code replication. Furthermore, because of this modularity, it is easy to see that behaviour trees can be described at varying levels of detail, as we have done in this paper whilst transitioning from the general BT in Figure~\ref{fig:gbt} to the specific one in Figure~\ref{fig:sbt}.
Finally, subtrees can be inserted or removed without changing the overall structure of the tree.

With respect to versatility, we have demonstrated that several different sub tasks can be accomplished in order of priority by being encapsulated into a single BT. This encapsulation delegates between tasks, and ensures that the control system can flexibly handle switching between the variety of tasks needed to be subsequently satisfied.

Lastly, with respect to robustness, we have demonstrated that, by using the PPA behaviour structure \cite{colledanchise2018book}, as we do throughout this paper, we can always ensure that controls are being chosen with priority of goal fulfilment in mind. In the case of the BT in Figure~\ref{fig:sbt}, at the overall level behaviours are being executed to ultimately achieve mission finalisation, which requires the successful completion of all previous nodes in the main sequence.

\section*{Acknowledgment}
This  work  was  supported  by  Stiftelsen  for StrategiskForskning  
(SSF)  through  the  Swedish  Maritime Robotics Centre (SMaRC) (IRC15-0046).



\bibliographystyle{IEEEtran}
\bibliography{btReferences}

\end{document}